\documentclass[conference]{IEEEtran}
\IEEEoverridecommandlockouts
\usepackage{cite}
\usepackage{amsmath,amssymb,amsfonts}
\usepackage{graphicx}
\usepackage{textcomp}
\usepackage{xcolor}
\usepackage{array}
\usepackage{float}
\usepackage[keeplastbox]{flushend}
\usepackage[inline]{enumitem}
\usepackage[utf8]{inputenc}
\usepackage[english]{babel}
\usepackage{makeidx}
\usepackage{graphicx}
\usepackage[hidelinks]{hyperref}
\usepackage{amsmath}
\usepackage[inline]{enumitem}
\usepackage{multirow}
\usepackage{subcaption}
\usepackage{caption}
\usepackage{color}
\usepackage{booktabs}
\usepackage{siunitx}
\usepackage{cite}
\usepackage{algpseudocode}
\usepackage{array}
\usepackage{float}
\usepackage{todonotes}
\usepackage[keeplastbox]{flushend}
\usepackage[linesnumbered,ruled,vlined]{algorithm2e}
\usepackage{amsfonts}
\usepackage{threeparttable}
\usepackage{ragged2e}
\usepackage[normalem]{ulem}

\def\BibTeX{{\rm B\kern-.05em{\sc i\kern-.025em b}\kern-.08em
    T\kern-.1667em\lower.7ex\hbox{E}\kern-.125emX}}
\newcommand{\NW}[1]{ {\color{blue}{#1}} }

\begin{document}

\title{Exploiting the Transferability of Deep Learning Systems Across Multi-modal Retinal Scans for Extracting Retinopathy Lesions\\
\thanks{This work is supported by a research fund from Khalifa University: Ref: CIRA-2019-047.}
}

\author{\IEEEauthorblockN{Taimur Hassan$^{\dagger}$ $^{\diamond}$, Muhammad Usman Akram$^{\diamond}$, Naoufel Werghi$^{\dagger}$}
\IEEEauthorblockA{$^{\dagger}$Department of Electrical Engineering and Computer Sciences, Khalifa University, Abu Dhabi, United Arab Emirates.\\
$^{\diamond}$Department of Computer and Software Engineering, National University of Sciences and Technology, Islamabad, Pakistan.
}
}

\maketitle

\begin{abstract}
Retinal lesions play a vital role in the accurate classification of retinal abnormalities. Many researchers have proposed deep lesion-aware screening systems that analyze and grade the progression of retinopathy. However, to the best of our knowledge, no literature exploits the tendency of these systems to generalize across multiple scanner specifications and multi-modal imagery. Towards this end, this paper presents a detailed evaluation of semantic segmentation, scene parsing and hybrid deep learning systems for extracting the retinal lesions such as intra-retinal fluid, sub-retinal fluid, hard exudates, drusen, and other chorioretinal anomalies from fused fundus and optical coherence tomography (OCT) imagery. 
Furthermore, we present a novel strategy exploiting the transferability of these models across multiple retinal scanner specifications. A total of 363 fundus and 173,915 OCT scans from seven publicly available datasets were used in this research (from which 297 fundus and 59,593 OCT scans were used for testing purposes). Overall, a hybrid retinal analysis and grading network (RAGNet), backboned through ResNet\textsubscript{50}, stood first for extracting the retinal lesions, achieving a mean dice coefficient score of 0.822. Moreover, the complete source code and its documentation are released at \url{http://biomisa.org/index.php/downloads/}.
\end{abstract}

\section{Introduction}
\label{sec:intro}
\noindent Retinopathy or retinal diseases tend to damage the retina, which may result in a non-recoverable loss of vision or even blindness if not timely treated. Most of these diseases are associated with diabetes. However, they may also occur due to aging, uveitis, and cataract surgeries. The two common retinal diseases are macular edema (ME) and age-related macular degeneration (AMD). ME occurs due to fluid accumulation within the macula mostly due to the associated hyperglycemia, uveitis, and cataract surgeries. ME caused by diabetes is often termed as diabetic macular edema (DME) which is identified by examining the patient’s diabetic history and also by analyzing the retinal thickening (caused due to retinal fluid) or hard exudates (HE) within the one-disc diameter of the center of the macula (containing a small pit known as fovea) \cite{_2}.
\begin{figure}[t]
\includegraphics[width=\linewidth]{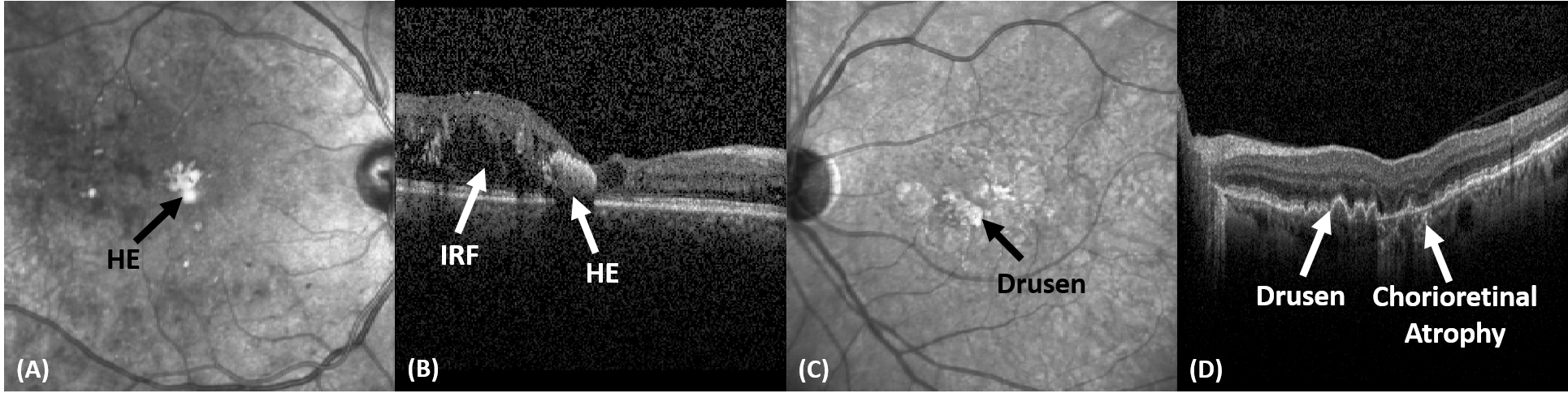}
\caption{ \small Retinal lesions in fundus and OCT scans of Ci-CSME (A, B) and dry AMD pathology (C, D).}
\centering
\end{figure}
Early Treatment Diabetic Retinopathy Studies (ETDRS) classified clinically significant ME as having: 1) either the thickening within 500$\mathrm{\mu}$m of the macular center; 2) HE along with the adjacent thickening within the macular center of 500$\mathrm{\mu}$m; or 3) retinal thickening regions of one (or more) disc diameter in which some part of them are within the one-disc diameter of the macular center \cite{_3}. 
But, with the advent of new imaging techniques such as optical coherence tomography (OCT), the classification of DME is redefined as centrally involved clinical significant macular edema (Ci-CSME) if the presence of retinal thickening, due to retinal fluid or hard exudates, is discovered within the central sub-field zone of the macula (having 1mm or greater diameter). Otherwise, DME is classified as non-centrally involved \cite{_4}. 
AMD is another retinal syndrome mostly found in elder people. It is typically classified into two stages i.e. non-neovascular AMD and the neovascular AMD. Non-neovascular AMD is the “dry” form of AMD in which small, medium or large-sized drusen can be observed. With the increasing disease severity, abnormal blood vessels intervene retina leading to chorioretinal anomalies such as fibrotic scars and choroidal neovascular membranes (CNVM). In such a case, AMD is classified as wet or neovascular AMD. Fig. 1 shows some of the fundus and OCT scans showing retinal lesions at different stages of AMD and DME.
\section{Related Work}
\noindent In the literature, a large body of solutions assessing retinal regions employed features extraction techniques coupled with classical machine learning (ML) tools. The majority of these methods are validated on a limited number of scans and thus exhibited feeble reproducibility. More recently, with the advent of deep learning, a wide variety of end-to-end approaches, operating on more massive datasets, have been proposed.

\noindent\textbf{Traditional Approaches:} Fundus imagery has been the modality of choice for examining the retinal pathology for a while \cite{_6} and is still used as a secondary examination technique in analyzing the complex retinal pathologies. But, with the advent of OCT, most of the solutions for retinal image analysis have migrated towards this new modality due to its ability to present objective visualization of retinal abnormalities in early stages. Chiu et al. \cite{_7} developed a kernel regression with graph theory and dynamic programming (KR+GTDP) scheme to extract retinal layers and retinal fluid from DME affected scans \cite{_7}. In \cite{_8}, a Random Forest-based framework was proposed for the automated extraction of retinal layers and fluid from scans affected by central serous retinopathy (CSR).  Wilkins et al. \cite{_9} presented an automated method for the extraction of intra-retinal cystoid fluid using OCT images. Vidal et al. \cite{_10} used a linear discriminant classifier, support vector machines, and a Parzen window for the identification of intra-retinal fluid (IRF). Apart from this, we have also proposed several methods for extracting retinal layers, retinal fluid, and for classifying retinopathy using traditional ML techniques \cite{_11,_12,_13,_14,_15}.
 
\noindent\textbf{Deep Learning Methods:} Many researchers have applied deep learning for the extraction of retinal layers \cite{_16} and retinal lesions such as IRF \cite{_18}, sub-retinal fluid (SRF) \cite{_19} and HE \cite{_23}.  Seebock et al. \cite{_21} proposed a Bayesian UNet based framework for recognizing different anomalies within the retinal pathologies. Fang et al. \cite{_22} developed a lesion-aware convolutional neural network (LACNN) model for the accurate classification of DME, choroidal neovascularization, drusen (AMD) and normal pathologies. LACNN is composed of a lesion detection network (LDN) and lesion-attention module where LDN first generates a soft attention map to weight the lesion-aware features extracted from the lesion-attention module and then these features are used for the accurate classification of retinal pathologies. Apart from this, we have recently proposed a hybrid retinal analysis and grading architecture (RAGNet) \cite{_23} that utilizes a single feature extraction model for retinal lesions segmentation, lesion-aware classification and severity grading of retinopathy based on OCT images.

\section{Contributions}
\noindent In this paper, we present a thorough evaluation of deep learning models for the extraction of IRF, SRF, HE, drusen, and other chorioretinal anomalies such as fibrotic scars and CNVM from multi-modal retinal images. Furthermore, we exploited the transferability of these models for retinal lesions extraction across multi-modal imagery. To the best of our knowledge, there is no literature available to date providing a thorough transferability analysis of encoder-decoder, fully convolutional, scene parsing, and hybrid deep learning systems for extracting this multitude of lesions in one go from multi-modal retinal imagery. Subsequently, the main contributions of this paper are:
\noindent\begin{itemize}[leftmargin=*]
  \item A first comprehensive evaluation of semantic segmentation, scene parsing and hybrid deep learning systems such as RAGNet \cite{_23}, PSPNet \cite{_24}, SegNet \cite{_25}, UNet \cite{_26}, and FCN (8s and 32s) \cite{_27} for extracting multiple lesions from multi-modal retinal imagery.
  \item A comprehensive study encompassing seven publicly available datasets, and five different retinal pathologies represented in a total of 363 fundus and 173,915 OCT scans from which 297 fundus and 59,593 OCT scans were used for testing purposes.
  \item A detailed exploration of the transferability of these models across multiple scanner specifications.
\end{itemize}

\section{Proposed Approach}
\noindent We propose a novel study to analyze the transferability of the state-of-the-art deep learning frameworks across fused fundus and OCT imagery for extracting multiple retinal lesions in one go. The models which we considered are as follows:

\noindent\textit{\textbf{RAGNet}}: is a hybrid convolutional network that can perform pixel-level segmentation and scan-level classification at the same time \cite{_23}. The uniqueness in the RAGNet architecture is that it uses the same feature extractor for the classification and segmentation purposes. So, if the problem demands segmentation and classification from the same image based upon similar features, then RAGNet would be an ideal choice rather than using two separate models \cite{_23}. Here, we have only used RAGNet segmentation unit since we are focusing on the retinal lesions segmentation.

\noindent\textit{\textbf{PSPNet}}: is a state-of-the-art scene parsing network that contains a pyramid pooling module to generate four pyramids of feature maps representing coarser to finer details to minimize the loss of global scene context while generating the latent representations \cite{_24}. The pooled outputs are then concatenated with the original feature maps to generate the final segmentation results. 

\noindent\textit{\textbf{SegNet}}: is an encoder-decoder network for semantic segmentation. The uniqueness in the SegNet model is that it uses pooling indices from the corresponding encoder block to up-sample the feature maps at the decoder end in a non-linear fashion. Afterward, the feature maps are convolved with trainable filters to remove their sparsity. Moreover, SegNet has a smaller number of trainable parameters due to which it is computationally more efficient. 

\noindent\textit{\textbf{UNet}}: is an auto-encoder inspired by FCN for semantic segmentation. The key feature of UNet is that it is fast and can generate good segmentation results with a small number of training samples because of its in-built data augmentation strategy \cite{_26}. UNet uses up-sampling instead of pooling operations and generates a large number of feature maps to mitigate the contextual information to the higher resolution layers \cite{_26}.

\noindent\textit{\textbf{FCN}}: is an end-to-end model proposed for semantic segmentation. FCN uses learned representation from the pre-trained models, fine-tune them, and generates finer pixel-level predictions in one go based upon up-sampling lower network layers with finer strides. In this study, we have utilized FCN-8 and FCN-32 (i.e. the finest and the coarsest version of FCN) for retinal lesions extraction. 

\noindent{We have applied these models for extracting retinal lesions from both fundus and OCT imagery. Here, we note that our study covers some of
the most complex and commonly occurring retinal pathologies including non-neovascular AMD, neovascular AMD, Ci-CSME, and non-Ci-CSME.  We also note that the related scans were collected using machines from different manufacturers and exhibit varying scan quality. 
 To make the comparison objective and highly reproducible, we have used publicly available datasets in our investigations. Furthermore, we have
tested the transferability of these models through an extensive cross-dataset validation.
The series of experiments we conducted in this work provide a reliable benchmark for assessing the robustness and generalization capacity of each model.} 
\section{Experimental Setup}
\noindent This section reports the detailed description of the datasets which have been used in this research. Furthermore, it contains implementation details and the performance metrics on which the models are evaluated:

\subsection{Datasets}
\noindent We have evaluated all the models on seven publicly available retinal image datasets where the ground truths for retinal lesions were acquired through the Armed Forces Institute of Ophthalmology, Rawalpindi Pakistan. The detailed summary of each dataset is presented below:

\noindent\textit{\textbf{Rabbani-I}} \cite{_28} is one of the few datasets which contains both OCT and fundus images of each subject reflecting AMD, DME, and normal pathologies. The dataset is acquired at Noor Eye Hospital, Tehran Iran and contains a total of 4,241 OCT and 148 fundus scans from 50 normal, 48 dry AMD, and 50 DME affected subjects. In this paper, we considered 37 fundus scans and 1,061 OCT scans for training and the rest for testing purposes.

\noindent\textit{\textbf{Rabbani-II}} \cite{_29} contains 12,800 OCT and 100 color fundus scans from both eyes of 50 healthy subjects. Since Rabbani-II only contain scans from the healthy subjects so it served as an excellent benchmark to test the false positive rate for all the models (indicating how many false positives each model generates).

\noindent\textit{\textbf{Duke-I}} \cite{_30} is one of the oldest retinal OCT datasets containing a total of 38,400 scans from which 26,900 scans reflect dry AMD and 11,500 scans show controlled (healthy) pathology. In the proposed study, a total of 300 scans have been used for training and 38,100 scans have been used for testing purposes. 

\noindent\textit{\textbf{Duke-II}} \cite{_7} has a total of 610 OCT images from 10 severe DME affected subjects. Moreover, the dataset contains highly detailed markings for retinal layers and fluid from two clinicians. In this paper, a total of 305 scans were used for training from the first five subjects, and the rest for testing.

\noindent\textit{\textbf{Duke-III}} \cite{_31} is another dataset from Duke University which we used in our research. The dataset contains 723 scans reflecting dry AMD pathologies, 1,101 scans reflecting DME pathology, and 1,407 scans showing normal retinal pathology. For the experimentations, we considered 3,048 scans for training and the rest 183 for testing.

\noindent\textit{\textbf{BIOMISA}} dataset \cite{_32} contains a total of 5,324 OCT (657 dry AMD, 2,195 ME, 904 normal, 407 wet AMD, and 1,161 CSR) and 115 fundus scans from 99 subjects (17 healthy, 31 ME, 8 dry AMD, 19 wet AMD, 24 CSR). In this study, a total of 1,299 OCT and 29 fundus images from the BIOMISA dataset were used for training and the rest for the evaluation purposes.

\noindent\textit{\textbf{Zhang}} dataset \cite{_33} contains 109,309 scans representing wet AMD (CNV), dry AMD (Drusen), DME, and healthy pathologies. The dataset also presents a clear separation of 108,309 scans for training while 1,000 scans for testing purposes which we followed in the experimentations as well.  




\subsection{Implementation Details}
\noindent All the models have been implemented using Keras, Python 3.7.4 on a machine having Intel 8\textsuperscript{th} generation Core i5, NVIDIA RTX 2080 GPU and 16 GB RAM where ResNet\textsubscript{50} was used as a backbone. Moreover, the optimizer used during the training was an adaptive learning rate method (ADADELTA) with a default learning and decay rate. The source code has been released at \url{http://biomisa.org/index.php/downloads/} for reproducibility.

\subsection{Evaluation Metrics}
\noindent In the proposed study, all the segmentation models have been evaluated using the following metrics:

\noindent\textit{\textbf{Mean Dice Coefficient}}: Dice coefficient ($\mathrm{D_C}$) computes the degree of similarity between the ground truth and the extracted results using following relation: $(\mathrm{D_C = \frac{2T_P}{2T_P + F_P + F_N}})$, where $\mathrm{T_P}$ indicates the true positives, $\mathrm{F_P}$ indicates the false positives and $\mathrm{F_N}$ indicates the false negatives. After computing $\mathrm{D_C}$ for each lesion class. The mean dice coefficient is computed for each network by taking an average of their $\mathrm{D_C}$ scores. 

\noindent\textit{\textbf{Mean Intersection-over-Union}}: The mean intersection-over-union (IoU) is computed by taking an average of IoU scores for each lesion class where the IoU scores are computed through $(\mathrm{IoU = \frac{T_P}{T_P + F_P + F_N}})$. 

\noindent\textit{\textbf{Recall, Precision and F-score}}: To further evaluate the models, we computed pixel-level recall $\mathrm{(T_{PR}=\frac{T_P}{T_P+F_N}})$, precision $\mathrm{(P_{PV}=\frac{T_P}{T_P+F_P}})$ and F-score $\mathrm{(F_1=\frac{2 x T_{PR} x P_{PV}}{T_{PR} + P_{PV}}})$.

\noindent\textit{\textbf{Qualitative Evaluations}}: The performances of all the models for lesions extraction have been also qualitatively evaluated through some visual examples.
\section{Results and Discussion}
\noindent The evaluation of segmentation models has been conducted on the combination of all seven datasets containing mixed OCT and fundus scans. In terms of $\mathrm{T_{PR}}$ and $\mathrm{F_1}$ as shown in Table I, RAGNet achieves 9.48\% and 3.36\% improvements as compared to UNet and PSPNet, respectively. However, in terms of precision, SegNet has a lead of 1.52\% as compared to PSPNet. This indicates that SegNet produces fewer false positives as compared to the rest of the models. For pixel-level comparison, we have excluded accuracy because it gives biased results towards a dominant-negative class i.e. the background. 
\begin{table}[t]
\footnotesize
\center
\caption{Performance evaluations in terms of pixel-level recall, precision and F-scores on combined dataset. Bold indicate the best performance.}
\begin{tabular}{llll}
\toprule
Network      & $\mathrm{T_{PR}}$ & $\mathrm{P_{PV}}$ & $\mathrm{F_1}$\\ \toprule
RAGNet       & \textbf{0.8547}   & 0.8606 & \textbf{0.8576}  \\ \hline
PSPNet    & 0.7540   & 0.9200 &  0.8287\\ \hline
SegNet        & 0.6388 & \textbf{0.9342} & 0.7587 \\ \hline
UNet & 0.7736 & 0.8842 & 0.8252   \\ \hline
FCN-8  & 0.6238 & 0.6165 & 0.6201 \\ \hline
FCN-32  & 0.4755 & 0.5611 & 0.5147 \\ \bottomrule
\end{tabular}
\label{tab:time}
\end{table}

\noindent{Tables 2 and 3 reports the performance of all the models for extracting retinal lesions in terms of mean $\mathrm{D_C}$ and mean IoU, respectively. From Table 2, it can be observed that RAGNet achieves the best mean $\mathrm{D_C}$ score of 0.822, leading PSPNet by 4.5\% and FCN-32 by 51.33\%. Moreover, in terms of mean IoU,} RAGNet also achieves the overall best performance (mean IoU: 0.710) showing a neat gap over its competitors for extracting IRF, SRF, and HE regions. In Table 3, the second-best performance is achieved by PSPNet that lags from RAGNet by 6.9\%. Also, we noticed that on fundus images UNet achieves optimal lesion extraction results with an overall performance comparable to that of PSPNet. Fig. 2 shows the qualitative results of all the models when trained on multi-modal images from all seven datasets at once, where we can notice the best overall performance of RAG-Net. It should be noticed that extracting lesions accurately from both modalities at once is quite challenging as their image features vary a lot.
\begin{figure}
\includegraphics[width=\linewidth]{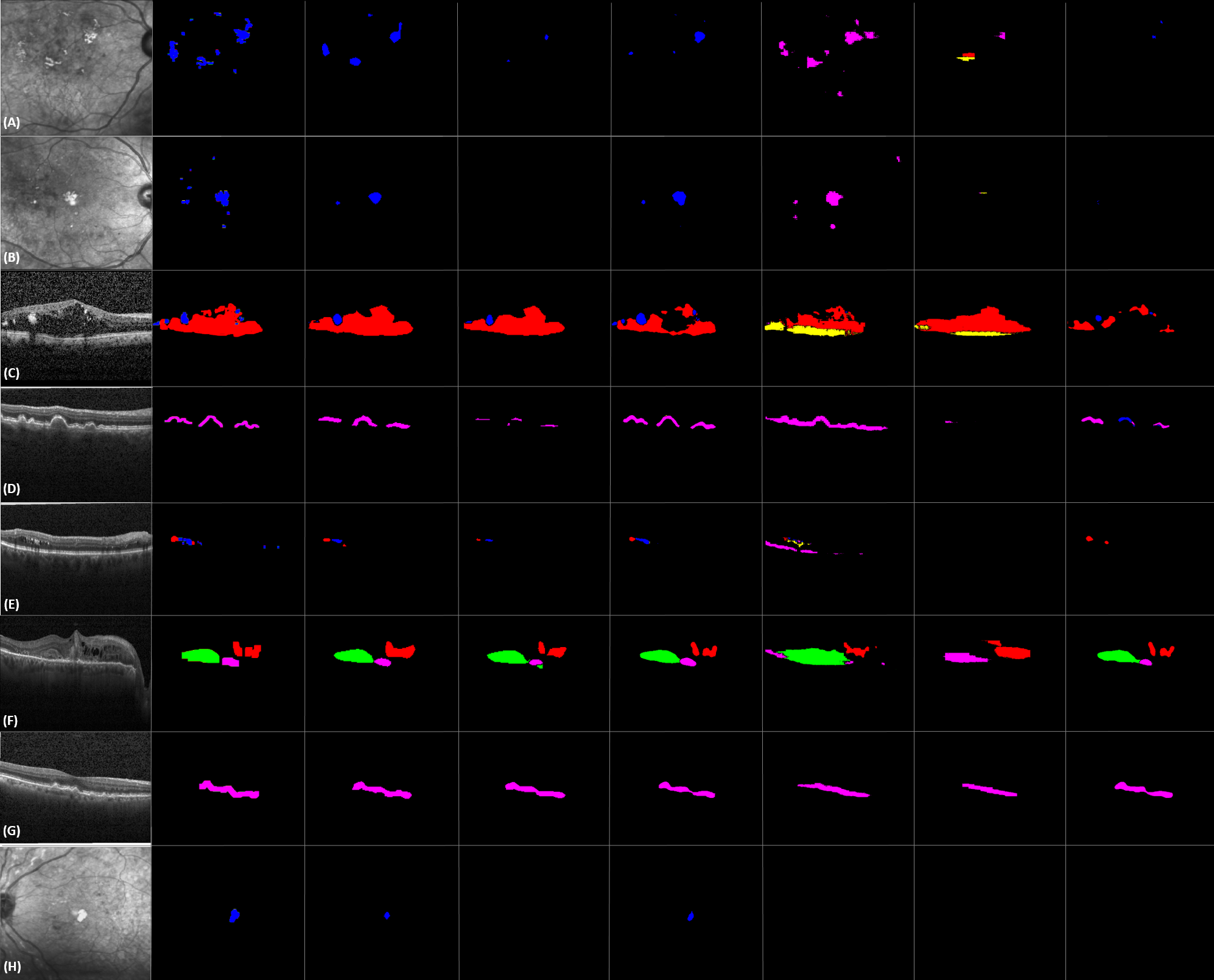}
\caption{ \small Comparison of retinal lesions extraction on combined dataset. (From left to right: original image, ground truth, RAGNet, PSPNet, UNet, FCN-8, FCN-32, SegNet). 
Blue, red, yellow, green, and pink indicate HE, IRF, SRF, CA, and drusen, respectively.}
\centering
\end{figure}
\begin{table}[htb]
\center
\caption{Performance evaluations of deep segmentation models for retinal lesions extraction in terms of $\mathrm{D_C}$. Bold indicates the overall best performance.}
\begin{tabular}{lllllll}
\toprule
Network & IRF & SRF & CA & HE & Drusen & Mean \\ \hline
RAGNet & \textbf{0.846} & \textbf{0.850} & 0.941 & \textbf{0.633} & 0.840 & \textbf{0.822}\\ \hline
SegNet & 0.810 & 0.610 &  0.886 & 0.373 & 0.695 & 0.675\\ \hline
PSPNet & 0.843 & 0.809 & \textbf{0.944} & 0.594 & 0.735 & 0.785\\ \hline
UNet & 0.816 & 0.757 & 0.878 & 0.581 & \textbf{0.864} & 0.779 \\ \hline
FCN-8 & 0.681 & 0.568 & 0.761 & 0.124 & 0.410 & 0.509 \\ \hline
FCN-32 & 0.651 & 0.434 & 0.638 & 0.032 & 0.243 & 0.400\\ \bottomrule
\end{tabular}
\label{tab:time}
\end{table}

\begin{table}[htb]
\footnotesize
\center
\caption{Performance evaluations of deep segmentation models for retinal lesions extraction in terms of IoU. Bold indicates the overall best performance.}
\begin{tabular}{lllllll}
\toprule
Network & IRF & SRF & CA & HE & Drusen & Mean \\ \toprule
RAGNet & \textbf{0.733} & \textbf{0.739} & 0.890 & \textbf{0.464} & 0.725 & \textbf{0.710}\\ \hline
SegNet & 0.681 & 0.439 &  0.796 & 0.229 & 0.533 & 0.535\\ \hline
PSPNet & 0.728 & 0.680 & \textbf{0.895} & 0.423 & 0.581 & 0.661\\ \hline
UNet & 0.689 & 0.609 & 0.783 & 0.409 & \textbf{0.761} & 0.650 \\ \hline
FCN-8 & 0.517 & 0.397 & 0.615 & 0.066 & 0.257 & 0.370 \\ \hline
FCN-32 & 0.482 & 0.277 & 0.468 & 0.016 & 0.138 & 0.276\\ \bottomrule
\end{tabular}
\label{tab:time}
\end{table}

\begin{table}[htb]
\footnotesize
\center
\caption{Transferability analysis (Training $\rightarrow$ Testing) for all models in terms of mean IoU. Bold and blue indicates the first and second-best performance, respectively. (Datasets name are coded as follows: R: Rabbani, D: Duke, Z: Zhang and B: BIOMISA). The rest of the abbreviations are RN: RAGNet, PN: PSPNet, SN: SegNet, UN: UNet, F-8: FCN-8, F-32: FCN-32.}
\begin{tabular}{lllllll}
\toprule
 & RN & PN & SN & UN & F-8 & F-32 \\ \toprule
R $\rightarrow$ D & \textbf{0.624} & \NW{0.589} & 0.414 & 0.574 & 0.281 & 0.170\\ \hline
D $\rightarrow$ R  & \textbf{0.649} & 0.601 &  0.426 & \NW{0.612} & 0.301 & 0.194\\ \hline
R $\rightarrow$ Z  & \textbf{0.657} & \NW{0.615} & 0.468 & 0.604 & 0.322 & 0.225\\ \hline
Z $\rightarrow$ R  & \textbf{0.663} & \NW{0.632} & 0.472 & 0.629 & 0.329 & 0.245 \\ \hline
B $\rightarrow$ R & \textbf{0.573} & \NW{0.542} & 0.389 & 0.534 & 0.236 & 0.144 \\ \hline
R $\rightarrow$ B & \textbf{0.554} & \NW{0.535} & 0.374 & 0.521 & 0.213 & 0.115\\ \hline
Z $\rightarrow$ D & \textbf{0.809} & \NW{0.752} & 0.613 & 0.734 & 0.476 & 0.342\\ \hline
D $\rightarrow$ Z & \textbf{0.794} & \NW{0.741} & 0.598 & 0.721 & 0.445 & 0.335\\ \hline
D $\rightarrow$ B & \textbf{0.582} & \NW{0.534} & 0.487 & 0.525 & 0.229 & 0.136\\ \hline
B $\rightarrow$ D & \textbf{0.571} & \NW{0.529} & 0.479 & 0.517 & 0.214 & 0.127\\ \hline
B $\rightarrow$ Z  & \textbf{0.564} & 0.513 & 0.465 & \NW{0.524} & 0.221 & 0.152\\ \hline
Z $\rightarrow$ B & \textbf{0.552} & 0.507 & 0.457 & \NW{0.512} & 0.235 & 0.178\\ \bottomrule
\end{tabular}
\label{tab:time}
\end{table}
\noindent{In the second series of experiments, we have conducted a transferability analysis to assess the generalization capabilities of all models. Here, we combined Duke-I, II, and III as one dataset (i.e. Duke) and Rabbani-I and Rabbani-II dataset as Rabbani to avoid redundant combinations as they have similar image features. We report the results in Table 4
where it can be observed that all the methods have shown good performance for Duke and Zhang dataset pairs and this is natural because both datasets are acquired through Spectralis, Heidelberg Inc. Moreover, RAGNet achieved the overall best performance as evident from Table 4, whereas PSPNet stood 2\textsuperscript{nd} best but its performance is comparable with UNet. In another experiment, we have used Rabbani-II dataset to test how many false positive does each model produce. Since Rabbani-II contains only healthy scans, so there are no actual lesions in this dataset. The best performance in this experiment is achieved by the RAGNet with a true negative $\mathrm{(T_N)}$ rate of 0.9999 indicating that it produces a minimum number of false lesions. Apart from this, the worse performance is achieved for FCN-32 ($\mathrm{T_N}$ rate: 0.9379). The worse performance of FCN-32 is even above 90\% because the ratio of $\mathrm{T_N}$ pixels and the $\mathrm{F_P}$ pixels is extremely high. The results for this experiment are available in the codebase package for the readers at http://biomisa.org/index.php/downloads/.}

\section{Conclusion and Future Research}
\noindent In this paper, we presented a thorough evaluation of semantic segmentation, scene parsing, and hybrid deep learning systems for extracting retinal lesions from fused fundus and OCT imagery. We also assessed the generalization capacity of each model through comprehensive cross-data validations where RAGNet, due to its robustness in retaining lesion contextual information during scan decomposition, produces superior results as compared to other models. Furthermore, the benchmarking performed in this work will be of great utility for both researchers and practitioners who want to employ deep learning models for lesion-aware grading of the retina.  In the future, we plan to extend and exploit this study for the extraction of the optic disc and retinal layers in the optic nerve head region for the glaucoma analysis.

\bibliographystyle{ieeetr}
\bibliography{main}

\begin{thebibliography}{10}

\bibitem{_2}
G.~M. Comers, ``Cystoid macular edema,'' in {\em Kellog Eye Center}.
\newblock Accessed: June 2019.

\bibitem{_3}
``Diabetic macular edema,'' in {\em EyeWiki}.
\newblock Accessed: November 4th, 2019.

\bibitem{_4}
N.~Relhan {\em et~al.}, ``The early treatment diabetic retinopathy study
  historical review and relevance to today’s management of diabetic macular
  edema,'' in {\em Current Opinion in Ophthalmology}.
\newblock Wolters Kluwer, May 2017.

\bibitem{_5}
M.~U. Akram {\em et~al.}, ``An automated system for the grading of diabetic
  maculopathy in fundus images,'' in {\em 19th International Conference on
  Neural Information Processing}.
\newblock November 12th-15th, 2012.

\bibitem{_6}
T.~Hassan {\em et~al.}, ``Review of oct and fundus images for detection of
  macular edema,'' in {\em IEEE International Conference on Imaging Systems and
  Techniques (IST)}.
\newblock September, 2015.

\bibitem{_7}
S.~J. Chiu {\em et~al.}, ``Kernel regression based segmentation of optical
  coherence tomography images with diabetic macular edema,'' in {\em Biomedical
  Optics Express}.
\newblock Vol. 6, No. 4, April 2015.

\bibitem{_8}
D.~Xiang {\em et~al.}, ``Automatic retinal layer segmentation of oct images
  with central serous retinopathy,'' in {\em IEEE Journal of Biomedical and
  Health Informatics}.
\newblock Vol 23, No. 1, January 2019.

\bibitem{_9}
G.~R. Wilkins {\em et~al.}, ``Automated segmentation of intraretinal cystoid
  fluid in optical coherence tomography,'' in {\em IEEE Transactions on
  Biomedical Engineering}.
\newblock pp. 1109-1114, 2012.

\bibitem{_10}
P.~L. Vidal {\em et~al.}, ``Intraretinal fluid identification via enhanced maps
  using optical coherence tomography images,'' in {\em Biomedical Optics
  Express}.
\newblock October 2018.

\bibitem{_11}
S.~Khalid {\em et~al.}, ``Automated segmentation and quantification of drusen
  in fundus and optical coherence tomography images for detection of armd,'' in
  {\em Journal of Digital Imaging}.
\newblock December 2017.

\bibitem{_12}
S.~Khalid {\em et~al.}, ``Fully automated robust system to detect retinal
  edema, central serous chorioretinopathy, and age related macular degeneration
  from optical coherence tomography images,'' in {\em BioMed Research
  International}.
\newblock March 2017.

\bibitem{_13}
T.~Hassan {\em et~al.}, ``Automated segmentation of subretinal layers for the
  detection of macular edema,'' in {\em Applied Optics}.
\newblock 55, 454-461, 2016.

\bibitem{_14}
B.~Hassan {\em et~al.}, ``Structure tensor based automated detection of macular
  edema and central serous retinopathy using optical coherence tomography
  images,'' in {\em Journal of Optical Society of America A}.
\newblock 33, 455-463, 2016.

\bibitem{_15}
A.~M. Syed {\em et~al.}, ``Automated diagnosis of macular edema and central
  serous retinopathy through robust reconstruction of 3{D} retinal surfaces,''
  in {\em Computer Methods and Programs in Biomedicine}.
\newblock 137, 1-10, 2016.

\bibitem{_16}
L.~Fang {\em et~al.}, ``Automatic segmentation of nine retinal layer boundaries
  in oct images of non-exudative amd patients using deep learning and graph
  search,'' in {\em Biomedical Optics Express}.
\newblock Vol. 8, No. 5, May 2017.

\bibitem{_18}
A.~G. Roy {\em et~al.}, ``Re{L}ay{N}et: retinal layer and fluid segmentation of
  macular optical coherence tomography using fully convolutional networks,'' in
  {\em Biomedical Optics Express}.
\newblock Vol. 8, No. 8, 1 August 2017.

\bibitem{_19}
T.~Schlegl {\em et~al.}, ``Fully automated detection and quantification of
  macular fluid in oct using deep learning,'' in {\em Ophthalmology}.
\newblock Vol. 125, No. 4, April 2018.

\bibitem{_20}
B.~Hassan {\em et~al.}, ``Deep ensemble learning based objective grading of
  macular edema by extracting clinically significant findings from fused
  retinal imaging modalities,'' in {\em MDPI Sensors}.
\newblock July 2019.

\bibitem{_21}
P.~Seebock {\em et~al.}, ``Exploiting epistemic uncertainty of anatomy
  segmentation for anomaly detection in retinal oct,'' in {\em IEEE
  Transactions on Medical Imaging}.
\newblock May 2019.

\bibitem{_22}
L.~Fang {\em et~al.}, ``Attention to lesion: Lesion-aware convolutional neural
  network for retinal optical coherence tomography image classification,'' in
  {\em IEEE Transactions on Medical Imaging}.
\newblock August 2019.

\bibitem{_23}
T.~Hassan {\em et~al.}, ``{RAG-FW}: A hybrid convolutional framework for the
  automated extraction of retinal lesions and lesion-influenced grading of
  human retinal pathology,'' in {\em IEEE Journal of Biomedical and Health
  Informatics}.
\newblock March 2020.

\bibitem{_24}
H.~Zhao {\em et~al.}, ``Pyramid scene parsing network,'' in {\em IEEE CVPR}.
\newblock 2017.

\bibitem{_25}
V.~Badrinarayanan {\em et~al.}, ``Segnet: A deep convolutional encoder-decoder
  architecture for image segmentation,'' in {\em IEEE Transactions on Pattern
  Analysis and Machine Intelligence}.
\newblock December 2017.

\bibitem{_26}
O.~Ronneberger {\em et~al.}, ``U-net: Convolutional networks for biomedical
  image segmentation,'' in {\em MICCAI}.
\newblock 2015.

\bibitem{_27}
J.~Long {\em et~al.}, ``Fully convolutional networks for semantic
  segmentation,'' in {\em IEEE CVPR}.
\newblock 2015.

\bibitem{_28}
R.~Rasti {\em et~al.}, ``Macular oct classification using a multi-scale
  convolutional neural network ensemble,'' in {\em IEEE Transactions on Medical
  Imaging}.
\newblock vol. 37, no. 4, pp. 1024-1034, April 2018.

\bibitem{_29}
T.~Mahmudi {\em et~al.}, ``Comparison of macular octs in right andleft eyes of
  normal people,''
\newblock in Proc. SPIE, Medical Imaging, San Diego, California, United States
  Feb. 15-20, 2014.

\bibitem{_30}
S.~Farsiu {\em et~al.}, ``Quantitative classification of eyes with and without
  intermediate age-related macular degeneration using optical coherence
  tomography,'' in {\em Ophthalmology}.
\newblock 121(1), 162-172 January 2014.

\bibitem{_31}
P.~P. Srinivasan {\em et~al.}, ``Fully automated detection of diabetic macular
  edema and dry age-related macular degeneration from optical coherence
  tomography images,'' in {\em Biomedical Optics Express}.
\newblock Vol. 5, No. 10 | DOI:10.1364/BOE.5.0035 68, 12 Sep 2014.

\bibitem{_32}
T.~Hassan {\em et~al.}, ``{BIOMISA} {R}etinal {I}mage {D}atabase for {M}acular
  and {O}cular {S}yndromes,'' in {\em ICIAR-2018}.
\newblock Portugal, June 2018.

\bibitem{_33}
D.~Kermany {\em et~al.}, ``Identifying medical diagnoses and treatable diseases
  by image-based deep learning,'' {\em Cell}.
\newblock 172(5):1122-1131, 2018.

\end{thebibliography}

\end{document}